# Development of a Hindi Lemmatizer

Snigdha Paul[#1], Nisheeth Joshi[#2], Iti Mathur[#3]
#Apaji Institute, Banasthali University, Rajasthan, India
[1]snigdha.pal18@gmail.com
[2]nisheeth.joshi@rediffmail.com
[3]mathur_iti@rediffmail.com

*Abstract*— We live in a translingual society, in order to communicate with people from different parts of the world we need to have an expertise in their respective languages. Learning all these languages is not at all possible; therefore we need a mechanism which can do this task for us. Machine translators have emerged as a tool which can perform this task. In order to develop a machine translator we need to develop several different rules. The very first module that comes in machine translation pipeline is morphological analysis. Stemming and lemmatization comes under morphological analysis. In this paper we have created a lemmatizer which generates rules for removing the affixes along with the addition of rules for creating a proper root word.

## I. INTRODUCTION

Morphological analysis is one of the most important part of linguistic analysis where we study the structure of words. This analysis is used to segment the words into morphemes. For the analysis of text in any language morphological analyzer is one of the foremost step. When we talk about language the first thing that comes in our mind is a "Word." Language is vast and has a huge diversification in words. These words make up a language. So in order to have knowledge about any language we need to know about its word structure. Language like Hindi is morphologically rich and need a keen analysis of words so that we can acquire its meaning and grammatical information. Lemmatization is the most important part of morphological analysis. With lemmatization we come to know the root word. Further by applying morphological tactics all the features of a particular word is shown. Morphology shows all the features of a particular word. By this analysis we come to know the category, gender, number, case, etc. thus we get all the information about a word which tend us to understand the language.

Morphology has two main broad classes: inflectional morphology and derivational morphology. Inflectional morphology is the study of those words which when inflected does not change the class and is formed from the existing stem, for example - we have a word *खज़ाना* which when added with a suffix becomes *खज़ानची*. The derived word doesn't change its word class that is both *खज़ाना* and *खज़ानची* belong to noun class. Derivational morphology is the study of those words which when inflected changes the class and new word formation takes place from the existing stem, for example – the word *सजा* when suffixed with *वट* gives a derived word सजावट which belongs to a different class, that is the word सजा is verb but when suffixed, gives us noun (*सजावट*).Thus we have root words which are affixed with several morphemes. Generally when we look up a dictionary we get almost all the root words but their derivated forms are rarely found.

To overcome this problem we have morphological analyzer. The analyzer is a way to get the derived word from the root thereby giving its features. In this paper we aim to develop a tool for executing the inflectional analysis of Hindi by using rule based approach. In this approach, in order to obtain the suffix list the foremost thing that we did is the study of various Hindi words. After getting the suffix list we created rules. These rules also include the addition and deletion of characters to make the word a proper root or stem. Here we do not emphasize on the grammatical information of the words which are categorized as number, person and gender. Along with this analysis the rest of the paper includes linguistic background of Hindi. Approach that we have applied is, generation of suffixes and rules.

## II. RELATED WORK

A lot of research work has been done and is still going on for the development of a stemmer as well as lemmatizer. The first stemmer was developed by Julie Beth Lovins [12] in 1968. Later the stemmer was improved by Martin Porter [11] in July, 1980 for English language. The proposed algorithm is one of the most accepted methods for stemming where automatic removal of affixes is done from English words. The algorithm has been implemented as a program in BCPL. Much work has been done in developing the lemmatizer of English and other European languages. In contrast, very little work has been done for the development of lemmatization for Indian languages. A rule based approach proposed by Plisson et al. [10] is one of the most accepted lemmatizing algorithms. It is based on the word endings where the suffix should be removed or added to get the normalized form. It emphasizes on two word lemmatization algorithm which is based on if-then rules and the ripple down approach. The work proposed by Goyal et al. [1] focuses on the development of a morphological analyzer and generator. They aimed to develop a translation system especially from Hindi to Punjabi. Nikhil K V S [5] built a Hindi derivational analyzer using a specific tool. He used supervised approach by creating a SVM classifier. Jena et al. [6] proposed a morphological analyzer for Oriya language by using the paradigm approach. They classified nouns, adjectives and finite verbs of Oriya by using various paradigm tables. Anand Kumar et al. [4] developed an automatic system for the analysis of Tamil morphology. They used various methodologies, rule based approach and sequence labelling containing the non linear relationships of morphological features from the training data in a better way.

Chachoo et al. [3] used an extract tool named Extract v2.0 for the development of the orthographic component of Kashmiri Script. A method has been proposed by Majumder et al. [9] in which a clustering based approach is used for discovering the equivalent classes of root words. This algorithm was tested for two languages French and Bangla. A rule based approach for stemming in Hindi was proposed by Ramanathan & Rao [8]. The approach is based on stripping off suffixes by generating rules emphasizing on noun, adjective and verb inflections in Hindi. Bharti Akshar et al. [2] proposed the work on natural language processing where they gave a detailed study of morphology using paradigm approach. Gupta et al. [13] proposed unsupervised approach for stemming where they have used partial lemmatization along with some database. They aimed to improve unsupervised stemming by removing over-stemming problem. Mohd. Shahid Husain [14] developed a stemmer by using unsupervised approach. He used two different approaches frequency based and length based method for suffix stripping where he used Emille corpus for Urdu and Marathi languages.

### III. LINGUISTIC BACKGROUND OF HINDI

Morphemes play a major role in morphology. This is the major way in which morphologists investigate words, then internal structure and their formation. Morphology is broadly categorized into two parts: derivational morphology and inflectional morphology. Derivational morphology processes the words and form new lexemes from the existing ones. This is done by either adding or deleting affixes. For example – सच्चा + ई = सच्चाई. The class of the word is changed from adjective to noun. Similarly, in English we have words like *computer + ization = computerization*, where the class is changed from noun to adjective. Inflectional morphology processes the words by producing various inflections without changing the word class. For example – किताब + ें = किताबें where किताब is noun/singular while किताबें is noun/plural. The class remains same here. The root form of the words basically comes under noun and verb classes. This knowledge lead us to trace the paradigm approach. According to Smriti Singh and Vaijayanthi M Sarma [7], Hindi noun classification system shows only the number and case for morphological analysis. Number basically includes either singular or plural. By default we keep the number of a word as singular. Case on Hindi words is of two types – direct and oblique. Oblique words show the case as well as the number of the word. For example – लड़क – ा, लड़क – े, here ा shows singular number whereas े shows plural number. Similarly we also have some gender rules. In Hindi, words that end with the suffix ी are marked feminine whereas the words that end with suffix ा are marked as masculine. For example लड़का is masculine ending with ा while लड़की is feminine ending with ी. But there are many words that contradict this concept. For example – we have the word पानी (water) which is masculine although it is ending with ी. Similarly we have another word माला (garland) which is feminine even though ending with ा. There are also some words which has the suffix that cannot be removed. For example – let us consider the suffix ा, the words पिता, माता, बच्चा, कटोरा, नेता, and many more does not require stemming. Such words need to be maintained as it is and should be refrained from being stemmed. So we find that Hindi is a highly inflected language which needs the deep study of word structure and its formation.

### IV. PROPOSED WORK

In this paper we have discussed about the creation of a Hindi lemmatizer. Our approach is based on the key concept of optimization. Optimization includes both space and time, so our approach is based on these parameters. The lemmatizer that we discuss here mainly focuses on the time complexity. Typically a lemmatizer is built using *rule based approach*. In rule based approach along with the rules, knowledgebase is created for storing the grammatical features. Although the knowledgebase creation requires a large amount of memory, but in respect of time it gives us the best, accurate and fast result. The reason behind this fast retrieval is that, a very short time is taken to search the input word from the knowledgebase. A study has been conducted in Tamil which shows that Tamil words have infinite set of inflections but Hindi words have finite set of inflections which are quite easy to maintain in the knowledgebase. We have restricted our knowledgebase to commonly used words which do not contain the proper nouns like the names of person and place.

*A. Approach Used*

Although there are many approaches for performing lemmatization like supervised approach, rule based approach, unsupervised approach, but among these approaches rule based approach is one of the most acceptable approach. We have used rule based approach for extracting the suffixes. In rule based approach we have created many different rules for eliminating the suffixes. Since rule based approach totally works according to the rules therefore there is a less chance of error in obtaining the output. Rule based approach also optimizes the work by which we can get the result in a blink of an eye.

*B. Suffix Generation*

For the development of a lemmatizer we have gone through various words with their suffixes and examined the morphological changes. These suffixes and changes led to the development of specific rules. For example – If we take the word कमज़ोरी (weakness) then we find that the word is derived by adding ◌ी suffix. Similarly there are many other words with the same suffix. Some of them are shown in Table I and II

TABLE I

EXAMPLE OF DERIVED WORDS WITH SUFFIX ◌ी

| Root Word | Derived Word |
|---|---|
| कमज़ोर | कमज़ोरी |
| खुश | खुशी |
| गरम | गर्मी |
| गरीब | गरीबी |
| सर्द | सर्दी |

TABLE II

SOME MORE SUFFIXES

| Root Word | Derived Word | Suffix |
|---|---|---|
| लड़की | लड़किया | ि◌या |
| साफ़ | सफ़ाई | ई |
| गंभीर | गंभीरता | ता |
| सौदा | सौदागर | गर |
| कच्चा | कच्चापन | पन |
| पढ | पढ़ाई | ◌ाई |
| असली | असलियत | इयत |

Since the work has been done manually therefore this phase was quite time consuming. The suffixes were generated by processing a corpus of 20,000 sentences from which 55 lakh words were manually stemmed out of which 112 suffixes were derived.

*C. Rule Generation*

After the generation of suffix list we have developed rules. We have created 112 rules which are framed in such a way that the suffix gets removed from the input word and if required, addition of character or 'maatra' takes place. For example – let us take the suffix ो◌ं. Fig 1 shows the working of the rule for this suffix.

लड़कों – ो◌ं + ा = लड़का
सड़कों – ो◌ं = सड़क

Fig. 1  Working of a Rule

Table III illustrates this process.

TABLE III

WORDS SHOWING THE SUFFIX ो◌ं

| | | Rule application | |
|---|---|---|---|
| Word | Root | Extraction of suffix | Addition of character |
| लड़कों | लड़का | ो◌ं | ा |
| बालकों | बालक | ो◌ं | — |
| नागरिकों | नागरिक | ो◌ं | — |

In the above table on removing the suffix ो◌ं we get their respective root word. But the word लड़कों is an exception here because on removing the suffix ो◌ं we need to add ा to the last letter of the word to make it a genuine root word' लड़का.' Similarly there are many other rules for removing the suffix and if necessary addition of character may also take place. Similarly we also have some other rules, like the rule for extracting the suffix ि◌यों which is shown in Table IV.

TABLE IV

WORDS SHOWING THE SUFFIX ि◌यों

| | | Rule application | |
|---|---|---|---|
| Word | Root | Extraction of suffix | Addition of character |
| लड़कियों | लड़की | ि◌यों | ◌ी |
| कहानियों | कहानी | ि◌यों | ◌ी |
| कवियों | कवि | ि◌यों | ि◌ (exception) |
| चिड़ियों | चिड़िया | ि◌यों | ि◌या (exception) |

Since we know that in Hindi, when we remove the plural, we need to add ◌ी to the last letter of the word. This is the general grammar rule. Table 4 mentions the rule for the suffix ि◌यों in which we have created a general rule for removing the suffix and adding ◌ी to the word, but we have some exceptions here which include the addition of ि◌ instead of ◌ी. In the above table we have also shown an exception in the last word चिड़ियों where the root form is चिड़िया. This word contains two suffixes together which are ियों and ों. This becomes hard for the system as it finds difficulty in picking up the correct rule for the particular word. Similarly there are many more exceptions for which we have generated different rules. To overcome such problems we have built a database in which such exceptional words are kept. Although this work requires much time but for the sake of fast and accurate result this approach is applied. The rule is shown in Fig 2-

```
If (root) present in (knowledgebase)
{
        Fetch the root from the list
        Display;
}
else if (root) not present in (knowledgebase)
{
        If (source) ends with (suffix)
        {
                Substring the source
                Display the root;
        }
}
```

Fig. 2 Rule procedure

*D. Algorithmic Steps*

The input word is first checked in database. If the word exists in the database then it is displayed as output but if the word doesn't exist in the database then the rules are accessed for stripping out the suffix. The rules work by deleting the suffix from the input. After deletion, if the word provides a proper meaning then it is displayed as a result otherwise a particular character or matra is added to the stripped word to make it a proper meaningful word. The steps are shown in Fig 3.

1. Check input word in knowledgebase.
2. Display if exist.
3. Otherwise access the rules.
4. Generate suffix stripping rules
    i. Delete the suffix.
    ii. Delete & add characters.

Fig. 3 Algorithm

## V. EVALUATION

The system is evaluated for its accuracy where we gave 500 words for lemmatization. Among these 500 words 456 words were correctly lemmatized and 44 words were incorrect because they violated both the exceptional and general rules. Accuracy of the system was computed using the following equation-

$$Accuracy = \frac{Number\ of\ Correctly\ lemmatized\ words}{Total\ Number\ of\ Words} \times 100$$

Accuracy= 91%

Some of the input words are shown in Fig 4-

नज़रें, सड़कों, लड़की, लड़कियाँ, खुशी, भारतीयता, मजदूरी, मिठाई, बालिकाओं, निश्वस्नीय, गौरवांवित, सफलताओं, लड़कों, मंज़िलें, विदा, ज्यादा, पढ़ाई, कवियों, तिजोरियों, सतरंगी, आतंकियों, बुनाई, नकारात्मक, नेताओं, अपमानित, चिड़ियों, संशोधन, शक्तिशाली, शीलस्य.

Fig. 4 Snapshot of inputs

The output of some of these words are shown in Table V-

TABLE V
SEPARATED LEMMA AND SUFFIXES

| Lemma | Suffix |
|---|---|
| नज़र | ें |
| सड़क | ों |
| लड़की | - |
| खुश | ी |
| भारत | ीयता |
| मजदूर | ी |
| बालिका | ओं |
| विश्वास | नीय |
| सफल | ताओं |
| लड़का | ों |
| संशोध | न |
| तिजोरी | यों |
| लड़की | ियाँ |
| ज्यादा | - |

Some of the wrong output words are shown in Fig 5-

विवेचना, उद्योगपति, कष्टकारी, कष्टकार, आध्यात्मक, ज्ञानात्मक, कलाकृति, शांतिप्रियता, मिष्ठान, गुणवत्ता, गुणकारी, निरंतर, नकलची, निंदनीय, स्रजनात्मक, सौभाग्यशाली, स्वावलंबन, तमन्नाएं, घ्रणित, दयालु, चौकीदार, चमकीला, विधुतीकरण.

Fig. 5 Snapshot of errors

## VI. CONCLUSION

In this paper we have discussed the development of a lemmatizer for Hindi. The work uses the rule based approach by creating knowledgebase which contains all the Hindi words that are commonly used in day to day life. The approach also emphasized on time optimization problem rather than on space. Since nowadays space is not at all a big problem, therefore our approach aimed to optimize time and generate accurate result in a very short period. Our system gave 91% of accuracy.